\documentclass{article}

\usepackage[preprint]{neurips_2022}

\usepackage[utf8]{inputenc} % allow utf-8 input
\usepackage[T1]{fontenc}    % use 8-bit T1 fonts
\usepackage{hyperref}       % hyperlinks
\usepackage{url}            % simple URL typesetting
\usepackage{booktabs}       % professional-quality tables
\usepackage{amsfonts}       % blackboard math symbols
\usepackage{nicefrac}       % compact symbols for 1/2, etc.
\usepackage{microtype}      % microtypography
\usepackage{graphicx}
\usepackage{caption}
\usepackage[table]{xcolor}
\usepackage{natbib}
\usepackage{comment}
\usepackage{subfigure}
\usepackage{siunitx}
\usepackage{array}

\usepackage{amsmath}
\usepackage{amssymb}
\usepackage{mathtools}
\usepackage{amsthm}

\usepackage{microtype}
\usepackage{graphicx}
\usepackage{subfigure}
\usepackage{booktabs} % for professional tables
\usepackage{textcomp}

\usepackage{siunitx}
\usepackage{todonotes}

\renewcommand{\vec}[1]{\mathbf{#1}}
\newcommand\zu{\vec{z}_\mathrm{u}}
\newcommand\zr{\vec{z}_\mathrm{r}}
\newcommand\xr{\vec{x}_\mathrm{r}}
\newcommand\xu{\vec{x}_\mathrm{u}}
\newcommand\xc{\vec{x}_\mathrm{c}}
\newcommand\x{\vec{x}}

\DeclarePairedDelimiter\set\{\}

\title{Unintended memorisation of unique features in neural networks}

\author{
  John Hartley \\
  The University of Edinburgh \\
  \texttt{john.hartley@ed.ac.uk} \\
  \And
  Sotirios A. Tsaftaris \\
  The University of Edinburgh \& The Alan Turing Institute \\
  \texttt{s.tsaftaris@ed.ac.uk} \\
}
\begin{document}

\maketitle

\begin{abstract}
Neural networks pose a privacy risk due to their propensity to memorise and leak training data. We show that unique features occurring only once in training data are memorised by discriminative multi-layer perceptrons and convolutional neural networks trained on benchmark imaging datasets. We design our method for settings where sensitive training data is not available, for example medical imaging. Our setting knows the unique feature, but not the training data, model weights or the unique feature’s label. We develop a score estimating a model’s sensitivity to a unique feature by comparing the KL divergences of the model’s output distributions given modified out-of-distribution images. We find that typical strategies to prevent overfitting do not prevent unique feature memorisation. And that images containing a unique feature are highly influential, regardless of the influence the images’s other features. We also find a significant variation in memorisation with training seed. These results imply that neural networks pose a privacy risk to rarely occurring private information. This risk is more pronounced in healthcare applications since sensitive patient information can be memorised when it remains in training data due to an imperfect data sanitisation process.
\end{abstract}

\section{Introduction}

% DNNs memorise data
%Deep Neural Networks (DNNs) are a powerful tool for classifying images \citep{LeCun1989, Krizhevsky2012-alexnet, He2016-resnet, Huang2017-densenet}.

Deep Neural Networks (DNNs) memorise training labels \citep{Zhang2021-understanding-mem} whether the training data are real, noisy, or random, and whether the labels are shuffled or not \citep{Arplt2017-a-closer-look-at-memorization}. A recent work by \citet{feldman-shorttaleaboutalongtail, feldman-what-neural-networks} has established theoretically and empirically that DNNs can achieve close to optimal generalisation error in image classification tasks when they memorise examples which are predominantly rare and atypical from long-tailed data distributions. It has also been shown empirically for several benchmark datasets that memorised examples have a large influence on the network predictions for atypical but visually similar examples in the test set \citep{feldman-what-neural-networks}.

\begin{figure}[t]
    \centering
    \includegraphics[width=0.45\linewidth]{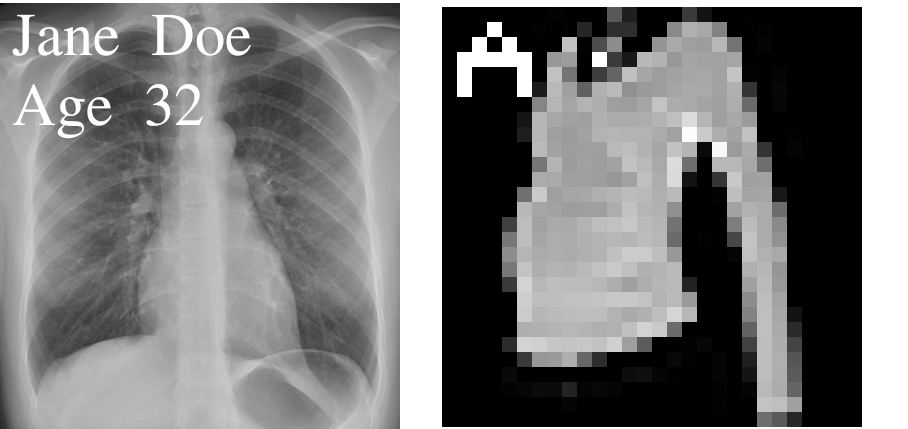}
    \caption{What happens if a training dataset of X-ray images mistakenly contains a \textit{single}  image that shows the name of a patient? Will this unique information be exploited (i.e.\ be memorised) by a neural network? Failure to remove sensitive information is possible as data sanitisation is not always perfect. We explore these questions by injecting simple visual features into images from known benchmark datasets (see example image on the right). We devise a memorisation score that assesses whether unintended \textit{unique feature memorisation} is possible. We explore whether architecture, regularisation,  stochasticity, and example influence matter. The short answer is \textit{Yes}. [This X-ray image is in the public domain \cite{wiki:Chest_radiograph}, and the patient name is fictitious.]}
    \label{fig:the-gist}
\end{figure}

%Nevertheless, memorisation of examples poses a privacy risk since information relating to the example is encoded directly in the weights of a neural network \cite{Golatkar2020, jegorova2021survey}. For example, an adversary could construct a readout function acting on the weights or network outputs to discover information about a given example \cite{shokri2017membership}. Data leakage is particularly problematic when datasets contain private information for which disclosure is limited. For example, DNNs used in healthcare may encode information about patients in their weights \citep{7163871, Zech2018-pneumonia-classification}, for which disclosure is limited in the EU by the General Data Protection Regulation (GDPR).

% Privacy risk of memorising features
%\sat{we need to chat about this para}

In this work we focus on the unintended memorisation of specific image \textit{features} that occur \textit{once in training data} as opposed to that of samples or training labels. We call this \textit{unique feature memorisation}.  While training examples have been shown to be memorised \citep{Zhang2021-understanding-mem} it is not clear whether an example is memorised in its entirety or whether specific features of the image are memorised. This distinction is important since private features in an image could be memorised, and leaked.  Memorisation of features poses a privacy risk as information derived from the feature is encoded directly in the network weights \citep{Golatkar2020, jegorova2021survey}. Consequently, as in a membership inference attack, an adversary could construct a readout function acting on the network's weights or outputs to discover information about a given feature \citep{shokri2017membership}. Such attacks exploit information about training data which is leaked by a model. In the feature setting this is called \textit{feature leakage} \citep{jegorova2021survey}.

Medical imaging offers a practical and realistic example of the risks posed by unique feature memorisation. For example, hospitals frequently employ sanitisation processes to remove patient names when they appear overlaid on X-ray films (see the example in Figure \ref{fig:the-gist}). These processes can fail, and occasionally an image with a patient's name will make it to a training dataset. A classifier trained on these data may misdiagnose other patients with the same name if those names also have not been removed. Or the unintended presence may lead to incomplete extraction of the correct discriminative features from the image \citep{DeGrave2021a}. Such a risk is similar to decision making based on spurious correlations, except that only a single spurious feature is present in the dataset \citep{7163871, Zech2018-pneumonia-classification, Geirhos2020-shortcut-learning, Idrissi2021}.

% Current privacy attacks
%We propose that models leak information about unique features they have learnt. Methods that exploit data leakage to uncover information about training data are called privacy attacks.

%In this work we show that models leak information about about unique features that they have learnt and that models memorise specific image features even when they occur only once in training data.

%One example is the membership inference attack (MIA) which finds whether an example is in the training set. This is achieved by exploiting a model's overconfidence on examples it has seen \citep{shokri2017membership, Sablayrolles2018, salem2018-ml-leaks, liu2020-have-you-forgotten, choquettechoo2021-label-only-mia}. Here instead we focus on the memorisation of unique features and not whole data. Adopting the terminology from \citet{jegorova2021survey} we will call this \textit{feature leakage}.

%Feature leakage occurs in large language models (LLM) due to unintended memorisation \citep{carlini2019secret, Carlini2020}. Unique sequences, such as credit card numbers or rarely occurring phrases, are random and therefore they are likely to be memorised by a model since they cannot be \textit{learnt} from similar patterns in the remainder of the training data.

We develop a score to detect unique feature memorisation in realistic settings, e.g.\ medical imaging, where access to data is limited by the data provider due to privacy concerns. We consider a setting where we know the task of the classifier and the unique feature but not the training data, the label of the image containing the unique feature, or the weights of the classifier.

In this setting feature memorisation cannot be evaluated using existing approaches. For example, In large language models (LLM) feature memorisation is quantified using the  exposure metric designed by \citet{carlini2019secret, Carlini2020}. In our setting we cannot use this metric. This is because we do not know the unique feature's label, and because unlike LLMs, the softmax layer of a discriminative model does not output the likelihood of the unique feature. Also, we cannot use leave-one-out methods to measure feature memorisation. This is because we cannot measure the difference in predictions from a model trained with and without the unique feature since we do not have access to the training data \citep{Koh2017, feldman-what-neural-networks}.

Our \textbf{main contributions} are summarised as follows:
\begin{itemize}
    \item We find that a \textbf{single} unique feature injected into benchmark datasets (MNIST, F-MNIST and CIFAR-10) can be memorised by a neural network before overfitting occurs.
    \item We offer a score for a black box setting to assess the memorisation of unique features in neural networks when there is no access to the training data and where only access to the model's softmax layer and the unique feature is allowed.
    \item We find that the risk of unique feature memorisation is not eliminated by adding explicit/implicit regularisation such as batch normalisation, dropout or data augmentation. On the contrary, we find that memorisation is more likely.
    \item We explore several potential factors that may influence unique feature memorisation including sample influence, stochasticity of the training process, and early stopping. We also explore where memorisation might happen and discuss mechanisms that may lead to this.
    
    %\item We find that sample-based influence measures are not sufficient to describe feature memorisation uniquely. We find the feature memorisation is not correlated with a sample based influence score and occurs randomly with the training seed. And that samples containing a unique feature have a high self-influence even when the other features do not.

    %\item We show that unique feature memorisation occurs early during training and before over-learning. Therefore it cannot be prevented by using techniques like early stopping.
    
    %\item We find that MLPs learn representations for unique features in their last hidden layer. And that these representations are not activated by random patches in the location of the unique feature in image space.
    
\end{itemize}

\section{Detecting feature memorisation}\label{sec:method}

% problem statement
%DNNs memorise training labels in image classification tasks. We propose that they \textit{also} independently memorise image features even when they occur \textit{once} in training data. We refer to a uniquely occurring feature as a \textit{unique feature}.

We present a score to approximately measure unique feature memorisation in a realistic worse-case scenario: an image pre-processor designed to remove features silently fails to remove a feature from a single training image. We consider a scenario where there is no access to the training data, or the unique feature’s label, and only access to the model's softmax layer. We know the unique feature’s data generating process, i.e.\ we know the content and location of the unique feature. (see Section \ref{sec:exp-setup}). We define the unique feature as a set of neighbouring pixels in a training image. We make the reasonable assumption that we know the domain of the classification problem e.g.\ X-ray images.

%\footnote{Leave-one-out memorisation measures do not apply as we do not have the training image without $\vec{z}_\mathrm{u}$.}

% Notation
\subsection{Notation}
We define a neural network image classification model $f(\vec{x};D_\mathrm{t})$, which maps an image $\vec{x}$ to a vector $\vec{y}$ where each element represents the conditional probability of the class label $y$ given the image $\vec{x}$, $D_\mathrm{t}=\{\vec{x}^i, y^i\}^N_{i=0}$ is the training data where $\vec{x}_p \in \mathbb{R}^{l \times l}$, and $y$ is the ground truth class label of $\vec{x}$. $D_\mathrm{t}$ may or may not contain a datum $\vec{x}_p$ having a unique feature $\vec{z}_\mathrm{u} \in \mathbb{R}^{m \times m}$ with $m < l$. Inspired by the work of \citet{carlini2019secret}, we call $\vec{x}_p$ a \textit{canary}. We define an additional random feature $\vec{z}_\mathrm{r} \sim U$, with the same dimensionality as $\vec{z}_\mathrm{u}$\footnote{There are several settings where some knowledge about $\zu$ exists. For example, in X-ray images a patient's name is written in a known location and typeface using a known alphabet.}. We make use of the KL divergence between two discrete probability distributions to measure the difference in the model's outputs, $D_{\mathrm{KL}}(P\vert\vert Q)= \mathbb{E}_{\mathrm{x}\sim P}\big[\log{\frac{P(x)}{Q(x)}}\big]$.

\subsection{Score for unique feature memorisation in a white box setting}\label{sec:greybox}

First we introduce a score for unique feature memorisation in a very simple theoretical setting where we have access to the training data $D_t$, and the unique feature's label $y$. This setting is of little practical use since access to the training data would be limited in settings where there is a risk disclosing sensitive information.  Nevertheless we provide it to concretely demonstrate the existence of unique feature memorisation. The score is given by $M_{w} = \mathbb{E}_{\x~\sim D_y} \big[\log(P(y|\xu)) - \log(P(y|\xr))\big]$, where $D_y$ is a subset of the training data $D_t$ containing examples with label $y$. We abuse the notation so that $\xu$, $\xr$ represent $\x$ with label $y$, injected with a unique/random feature patch respectively. Intuitively $M_{w}$ is the average difference in the label log-likelihoods between inferences on the unique/random feature IID samples. The scale of $M_w$ is straightforward to interpret  since the unique feature has a direct contribution to the model's prediction for $y$.

\subsection{Score for unique feature memorisation in a black box setting}\label{sec:m-score}

We now develop a memorisation score $M$ for practical settings where disclosure agreements prevent us from having access to the training dataset or its distribution.

% central premise
\subsubsection{Main idea}
To approximate the memorisation of $\zu$ we measure the sensitivity of $f$ to a set of image pairs which are clean, i.e.\ images not containing $\zu$ or $\zr$, vs.\ those containing $\vec{z}_r$ or $\vec{z}_{u}$. We hypothesise that if $f$ has memorised $\zu$ will be more sensitive to images containing $\zu$ than $\zr$. Any learning of $\zu$ which occurs must be memorisation since $\zu$ is unique and cannot be learnt from any other label structure in the training data.

\begin{figure}[t]
\centering     %%% not \center
\subfigure[$\xc$]{\label{fig:a}\includegraphics[width=0.12\linewidth]{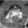}}\hfill
\subfigure[$\xu$]{\label{fig:b}\includegraphics[width=0.12\linewidth]{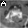}}\hfill
\subfigure[$\xr$]{\label{fig:c}\includegraphics[width=0.12\linewidth]{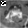}}\hfill
\subfigure[MNIST canary]{\label{fig:2d}\includegraphics[width=0.12\linewidth]{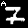}}
\hfill
\subfigure[F-MNIST canary]{\label{fig:2e}\includegraphics[width=0.12\linewidth]{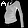}}
\hfill
\subfigure[CIFAR-10 canary ]{\label{fig:2f}\includegraphics[width=0.12\linewidth]{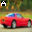}}
\caption{Example image pairs from greyscale CIFAR-10 used for inference on $f$ trained on MNIST or Fashion-MNIST, and canaries for testing unique feature memorisation. Each canary has an \textit{A} burnt into its top-left corner.}
\label{fig:inference-images}
\end{figure}

Given no access to the training dataset or its distribution we perform these inferences using an out-of-distribution (OOD) dataset. We construct three such datasets to probe $f$: $D_\mathrm{c}=\{\vec{x_c}^i\}^n_{i=0}$, $D_\mathrm{u}=\{\vec{x_u}^i\}^n_{i=0}$, $D_\mathrm{r}=\{\vec{x_r}^i\}^n_{i=0}$. $D_c$ is OOD to the training dataset, $D_t$. We assume that any dataset which is not the training set, $D_\mathrm{t}$, is OOD and can be used for inference. The specific distribution is not important since our method finds only the relative distances between model outputs from image pair inputs. $D_\mathrm{u}$ and $D_\mathrm{r}$ have the same samples as $D_c$ except that every image in $D_u$ also contains $\zu$ and every image in $D_r$ contains $\zr$, where $\zr$ is drawn randomly for every image. Figure \ref{fig:inference-images} shows examples of $\xc$, $\xu$, $\xr$.

\subsubsection{Model sensitivity to unique features in OOD data}

We use the KL divergence as a measure of the difference in sensitivity between ${\xc, \xu}$ and ${\xc, \xr}$ image pairs. We assume that  $f(\vec{x}_\mathrm{c})$, $f(\vec{x}_\mathrm{u})$, $f(\vec{x}_\mathrm{r})$ are valid probability distributions, and such that the KL divergences of any combination are also valid. We measure the network outputs of $f$ for every image pair in the OOD inference datasets $D_\mathrm{c}$, $D_\mathrm{u}$, $D_\mathrm{r}$ by calculating
%the following divergences

\noindent\begin{minipage}{.5\linewidth}
% KL divergences for the M score
\begin{align*}
    d_u^i = D_{\mathrm{KL}}\big(
    f(\vec{x}_\mathrm{c}^i)
    \vert\vert
    f(\vec{x}_\mathrm{u}^i)
    \big),
\end{align*}
\end{minipage}
\begin{minipage}{.5\linewidth}
\begin{align*}
    d_r^i = D_{\mathrm{KL}}\big(
    f(\vec{x}_\mathrm{c}^i)
    \vert\vert
    f(\vec{x}_\mathrm{r}^i)
    \big),
\end{align*}
\end{minipage}
where $d_u^i$, and $d_r^i$ measure the distances (divergences) in the predictions from $\xc^i$ to $\xu^i$ and $\xr^i$ where $i$ is a sample index in the datasets $D_u$, $D_c$ and $D_r$.  When the divergences $d_u^i$, and $d_r^i$ are zero there is no difference between the prediction on the clean image and the random or unique feature image. Whereas a network that is more sensitive to $\zu$ than any given $\zr$ will have $d_u^i > d_r^i$. Since $\zr$ is random it is possible that $\zr$ is already a feature learnt by the classifier, and thus $d_u^i < d_r^i$. However, we assume that the subspace of such features is far smaller than the randomness space of $\zr$, and thus most examples that we draw will not be features learnt by the classifier.

\subsubsection{The M score}

As the samples are OOD and random, we define our memorisation score for $\zu$ as
\begin{equation}
    M = \textrm{Average}(X_u) - \textrm{Average}(X_r)
\end{equation}
where $M>0$ corresponds to memorisation of the unique feature and the sets 

\noindent\begin{minipage}{.48\linewidth}
% KL divergences for the M score
\begin{align*}
    X_u = \set{D_{\mathrm{KL}}\big(
    f(\vec{x}_\mathrm{c}^i)
    \vert\vert
    f(\vec{x}_\mathrm{u}^i)
    \big) \big | \; 0\le i < n},
\end{align*}
\end{minipage}
\begin{minipage}{.48\linewidth}
\begin{align*}
    X_r = \set{D_{\mathrm{KL}}\big(
    f(\vec{x}_\mathrm{c}^i)
    \vert\vert
    f(\vec{x}_\mathrm{r}^i)
    \big)\big| \; 0 \le i < n}.
\end{align*}
\end{minipage}

We abuse the notation such that every $\xr^i$ is injected with a new random unique feature $\zr^i$. $\textrm{Average}(X_u)$ measures the $f$'s sensitivity to a unique feature and marginalises over the random effects of performing inferences on $f$ with OOD data \citep{shao2020understanding}. $\textrm{Average}(X_r)$ marginalises over the choice of random patch, and calibrates against random effects (e.g.\ boundary
interactions) of injecting any patch into OOD images.

\textit{Statistical significance.} Large values of $M$ correspond to greater memorisation since the signal from the unique feature is greater. We quantify the statistical significance of the $M$ score using a one-tailed t-test with an alternative hypothesis that the population mean of $X_u$ is greater than that of $X_r$.

%$\textrm{Mean}(X_u) > \textrm{Mean}(X_r)$.

%Therefore, to mitigate the similarity of random features to those learnt by the classifier and the robustness issues, we measure the sensitivity of the network by averaging over samples of $\zr$ and $\xc$. 

%
% assumptions
%where $P$ is the data distribution of $D_c$. We assume that drawing $\xc$ also draws the image pair $\xu$ with the unique feature. We approximate $X_u$ and $X_r$ by sampling from a benchmark dataset that is not the training dataset, and by $\vec{z}_r\sim U$ for every image in the OOD dataset.

%For memorisation tests on MNIST and Fashion-MNIST we use a greyscale version of CIFAR-10 for inference, and for tests on CIFAR-10 we use a 3-channel version of the MNIST dataset. 

% metric and uncertainty quantification.
%\noindent \textbf{The score $M$} We define our memorisation score for $\zu$ as $M = \textrm{Average}(X_u) - \textrm{Average}(X_r)$, where $M>0$ corresponds to memorisation of the unique feature. Large values of $M$ correspond to greater memorisation. We quantify the statistical significance of $M$ a using a one-tailed t-test for $\textrm{Mean}(X_u) > \textrm{Mean}(X_r)$.

\subsection{Experimental setup to measure feature memorisation}\label{sec:exp-setup}

\textit{Constructing artificial training datasets}. We evaluate $M$ by creating an artificial dataset $D_t$ that imitates the action of the pre-processor on the training data. To do this we augment a single training image in the dataset with a tiny $5\times5$ patch of a letter character $1\times1$ pixels from the top left corner of the image. Examples of these canaries injected with the unique feature \textit{A} are shown in Figures \ref{fig:2d}, \ref{fig:2e}, \ref{fig:2f}. Subsequently we train $f$ on $D_t$ and measure the memorisation of $\zu$ using the $M$ score.

\textit{Training models}. Overparameterised neural networks memorise random training labels when they are trained indefinitely \citep{Zhang2021-understanding-mem}. However, LLMs also memorise unique phrases even before overfitting in an average sense occurs \citep{carlini2019secret}. We call these models \textit{well-trained}. We suggest that unique features are also memorised by image classification models even when these models are well-trained. We test this hypothesis using early stopping when training with canaries. See Section \ref{sec:models-and-training} (in the supplemental) for the training strategy.

\textit{Selecting high self-influence examples as canaries}. Memorisation can occur when DNNs generalise from examples that are mislabelled or belong to sub-populations/long-tails within classes to samples in the test set \citep{feldman-shorttaleaboutalongtail, feldman-what-neural-networks}. Similarly, research into long-tailed learning establishes that models have a worst-group accuracy as a result of class imbalances or spurious correlations in datasets \citep{Liu2020a, zhang2021deep, Liu2021b}. Such examples can be described using self-influence: the degree to which learning on an example affects the models prediction on itself. We can estimate this  using influence functions, or proxy functions to influence functions \citep{Koh2017, pmlr-v80-katharopoulos18a, carlini2019prototypical, Ghorbani2019, Toneva2019, feldman-what-neural-networks, Garima2020, Guo2020-fastif, Baldock2021, Harutyunyan2021, Jiang2021}.

We would like to determine whether sample-based influence scores are related to feature-based memorisation \citep{Krueger2019}. We investigate two aspects. First that unique features injected on such examples are more likely to be memorised. Second that examples which contain the unique feature are always high influence.

 %We want to investigate these on the highest and lowest self-influence Cs. However, 
 %Since c
 Computing self-influence for every example in a dataset is computationally expensive since we must train a model for every example. Instead we use an approximation, TracIn, to estimate self-influence in a single training run \citep{Garima2020}. Then we study unique feature memorisation on 30 canaries with the highest and lowest self-influence. See Section \ref{sec:models-and-training} for details on computing TracIn.

\section{Experiments}\label{sec:experiments}

We present a series of experiments demonstrating memorisation of unique features in benchmark datasets for several image classification architectures. Supplementary experiments are provided in Section \ref{sec:add-exp}. Our experiments can be replicated with code from an online repository (see Section \ref{sec:models-and-training}).

%\textcolor{blue}{Our primary aim is to show that MLPs and CNNs memorise unique features occurring once in training data. In Section \ref{sec:whitebox-m} we show that the memorisation score detects memorisation and lack of memorisation in a theoretical whitebox setting for with MLPs on MNIST \citep{lecun2012mnist} on a CNN on CIFAR-10 \citep{krizhevsky2009-cifar10}. Next, we show that memorisation is still detected in a practical setting without access to training data or the canaries' label (Section \ref{sec:blackbox-m}), and that regularisation strategies do not prevent memorisation. In the remaining sections we aim to characterise unique feature memorisation by: assessing the effect of random training seed on memorisation given specific canaries (Section \ref{sec:stochastic-m}), finding features in MLPs which correspond to the unique feature (Section \ref{sec:activation-m}), profiling memorisation during training (Section \ref{sec: profile-m}), and quantifying the effect of sample based self-influence on feature memorisation (Section \ref{sec:influence-m})}.

\subsection{Neural networks memorise unique features}\label{sec:whitebox-m}

We begin with a simple white box setting to show that unique features occurring \textit{once} in training data are memorised by MLPs and CNNs. We measure $M_w$ for 100 randomly selected canaries with one unique instance per dataset in the white box setting (see Section \ref{sec:exp-setup}). Datasets and training details are provided in Section \ref{sec:models-and-training}. The top-5 $M_w$ canaries are shown in Table \ref{table:memorisation-whitebox}. Per-canary results are provided in Section \ref{sec:add-exp}. We find that injecting the unique feature into the training images significantly increases the average confidence of the predictions on the canaries' label in comparison with the random feature. For some canaries $M_w<0$. This shows the unique feature is not always memorised. We will see in Section \ref{sec:discussion} that the randomness of the learning algorithm may explain this.

\begin{table}
    \centering
    \caption{Memorisation scores ($M_w$) of unique features in MLP-1 and CNN-2 trained on MNIST and CIFAR-10 in a white box setting. Results in bold correspond to test statistics with p-values $<0.05$.\\}
    \label{table:memorisation-whitebox}
    \begin{small}
    \begin{tabular}{l c c r r r}
    \toprule
    IMAGE ID &  $D_t$ & MODEL & $\textrm{Average}(P(y|\vec{x}_u))$ & $\textrm{Average}(P(y|\vec{x}_r))$ &            $M_w$ \\
    \midrule
       16277 &    MNIST & MLP-1 &                               .50 &                               .44 & \textbf{.11} \\
       35730 &    MNIST & MLP-1 &                               .52 &                               .46 & \textbf{.11} \\
       48407 &    MNIST & MLP-1 &                               .48 &                               .43 & \textbf{.11} \\
       51668 &    MNIST & MLP-1 &                               .42 &                               .38 & \textbf{.10} \\
       16539 &    MNIST & MLP-1 &                               .41 &                               .37 & \textbf{.10} \\
       45571 & CIFAR-10 & CNN-2 &                               .72 &                               .66 & \textbf{.09} \\
       36230 & CIFAR-10 & CNN-2 &                               .58 &                               .53 & \textbf{.09} \\
       16926 & CIFAR-10 & CNN-2 &                               .64 &                               .59 & \textbf{.09} \\
       35437 & CIFAR-10 & CNN-2 &                               .61 &                               .56 & \textbf{.09} \\
       41457 & CIFAR-10 & CNN-2 &                               .53 &                               .49 & \textbf{.08} \\
    \bottomrule
    \end{tabular}
    \end{small}
    \end{table}

\begin{table}
    \centering
    \caption{Memorisation ($M$) scores of unique features in MLP-1, CNN-1, CNN-2, and a DenseNet trained on MNIST, F-MNIST and CIFAR-10. Results in bold correspond to test statistics with p-values $<0.05$. The average $M$ score for memorised canaries are highlighted in grey.\\}
    \label{table:memorisation}
    \begin{small}
    \begin{minipage}[t]{0.45\linewidth}\centering
        \begin{tabular}[t]{l l l l}
        \toprule
            ID &  $D_t$ &    MODEL &            $M$ \\
        \midrule
        27225 &   MNIST &  MLP-1 & \textbf{.0038} \\
        6885 &   MNIST &   MLP-1 & \textbf{.0021} \\
       27155 &   MNIST &   MLP-1 & \textbf{.0017} \\
       %11708 &   MNIST &   MLP-1 & \textbf{.0012} \\
       %8898 &   MNIST &   MLP-1 & \textbf{.0007} \\
       37251 & F-MNIST &   MLP-1 &   \textbf{.03} \\
        2731 & F-MNIST &   MLP-1 &  \textbf{.021} \\
        2181 & F-MNIST &   MLP-1 &  \textbf{.019} \\
        %3694 & F-MNIST &   MLP-1 &  \textbf{.019} \\
       %16002 & F-MNIST &   MLP-1 &  \textbf{.013} \\
        %\bottomrule
        %\end{tabular}
    %\end{minipage}
    %\begin{minipage}[t]{0.32\linewidth}\centering
    %    \begin{tabular}[t]{llll}
    %    \toprule
        
        51508 &    MNIST &    CNN-1 & \textbf{.073} \\
        14873 &    MNIST &    CNN-1 & \textbf{.035} \\
         7080 &    MNIST &    CNN-1 & \textbf{.021} \\
        %43454 &    MNIST &    CNN-1 &  \textbf{.02} \\
        %47034 &    MNIST &    CNN-1 &  \textbf{.02} \\
        59677 &  F-MNIST &    CNN-1 & \textbf{.085} \\
        23711 &  F-MNIST &    CNN-1 & \textbf{.068} \\
        15748 &  F-MNIST &    CNN-1 & \textbf{.059} \\
        %12168 &  F-MNIST &    CNN-1 & \textbf{.058} \\
        %10477 &  F-MNIST &    CNN-1 &  \textbf{.05} \\
        \bottomrule
        \end{tabular}
    \end{minipage}
    \begin{minipage}[t]{0.45\linewidth}\centering
        \rowcolors{8}{gray!20}{gray!20}
        \begin{tabular}[t]{l l l l}
        \toprule
            ID &  $D_t$ &    MODEL &            $M$ \\
        \midrule
        23308 & CIFAR-10 &    CNN-2 & \textbf{.058} \\
         9461 & CIFAR-10 &    CNN-2 & \textbf{.023} \\
         7371 & CIFAR-10 &    CNN-2 & \textbf{.018} \\
        %35174 & CIFAR-10 &    CNN-2 & \textbf{.017} \\
        %15726 & CIFAR-10 &    CNN-2 & \textbf{.013} \\
        32574 & CIFAR-10 & DenseNet &  \textbf{.72} \\
          772 & CIFAR-10 & DenseNet &  \textbf{.25} \\
         8022 & CIFAR-10 & DenseNet &  \textbf{.18} \\
         %\midrule
         AVG. &    MNIST &       MLP-1 &      {.0013} \\
         AVG. &    F-MNIST &     MLP-1 &      {.0093} \\
         AVG. &    MNIST &       CNN-1 &    {.012} \\
         AVG. &    F-MNIST &     CNN-1 &    {.028} \\
         AVG. &    CIFAR-10 &    CNN-2 &    {.018} \\
         AVG. &    CIFAR-10 &    DenseNet & {.38} \\
        \bottomrule
        \end{tabular}
    \end{minipage}
    \end{small}
    \end{table}

\subsubsection{Memorisation of unique features in MLPs}\label{result:1}

Now we proceed with the more practical black box setting where we do not know the training data or the canaries' label. We measure $M$ for the 15 highest and the 15 lowest self-influence training examples. See Section \ref{sec:models-and-training} for additional training details. Table \ref{table:memorisation} shows the top-3 $M$ canaries, and the average $M$ score over canaries with $M>0$ per dataset. %Averaging allows a comparison of memorisation across different approaches since it isolates instance-based variation. 
We will use the average memorisation score as a comparison between model architectures where the choice of canaries varies. The variation occurs because canaries are chosen by the self-influence score which depends on the model architecture. We report on the effect of self-influence on $M$ in Section \ref{sec:discussion}.

\subsubsection{Memorisation of unique features in CNNs}\label{result:2}

Here we investigate memorisation in CNNs in a black-box setting. Table \ref{table:memorisation} shows the top-3 $M$ canaries and the averaged $M$ score for canaries with $M>0$. Average memorisation appears greater in a CNN as opposed to an MLP architecture for MNIST ($0.012 > 0.0013$) and Fashion-MNIST ($0.028 > 0.0093$). In fact, average memorisation in CNNs is over an order of magnitude greater than MNIST. It also seems that the average memorisation is greater for CIFAR-10 in DenseNet than in CNN-2 ($0.38 > 0.018$), however, the sample size for the DenseNet results is small. Similarly to \citet{feldman-what-neural-networks}, networks with higher validation accuracies appear to show greater memorisation (DenseNet and CNN-2 have a validation accuracies of $\approx$92\% and  $\approx$72\% respectively).

%\jh{which canaries should we take the average over? top-5? canaries that have a positive memorisation score? all canaries?}

\subsection{Effects of explicit and implicit regularisation on memorisation}\label{result:3}

We now explore how explicit and implicit regularisation strategies affect unique feature memorisation. We consider three regularisation strategies: dropout, data augmentation, and batch normalisation, applied in various combinations. See Section \ref{sec:models-and-training} for details on the regularisers.

Table \ref{table:regularisation} shows the average memorisation score for canaries where $M>0$. The results, taking MNIST for example, illustrate that almost all values are higher than the average of 0.0013 obtained from Table \ref{table:memorisation} without any regularisation. We find that there is no statistically significant difference in the average memorisation scores when the regularisers are used to train MLP-1 on F-MNIST. Canaries in CIFAR-10 have greater memorisation scores than in the non-regularised models shown in Table \ref{table:memorisation}. These results extend to the level of features, findings that have been made previously for the memorisation of whole training examples \citep{Zhang2021-understanding-mem}.

Data augmentation increases memorisation. The average memorisation increases from 0.018 to 0.13 for CNN-2 trained on CIFAR-10 with data augmentation. We suggest this is because at every epoch a different perturbed canary is used to train the model. This increase in the number of effective canaries increases the spurious correlation between the unique feature and label and thus enables easier learning of the feature.

\begin{table}[t]
    \centering
    \caption{Average memorisation ($M$) scores over canaries with $M>0$ for unique features for models with explicit and implicit regularisers, such as Dropout, Data Augmentation, and Batch Normalisation.\\}
    \label{table:regularisation}
    \begin{small}
    \begin{tabular}{l c l r}
    \toprule
    $D_t$ & MODEL &       REGULARISATION &             $M$ \\
    \midrule
           MNIST &   MLP-1 &       Dropout &            .02 \\
           MNIST &   MLP-1 &       Data Augmentation &          .0033 \\
           MNIST &   MLP-1 & Dropout \& Data Augmentation &          .0013 \\
           MNIST &   MLP-1 &       Batch Normalisation &           .039 \\
         F-MNIST &   MLP-1 &       Dropout &           .025 \\
         F-MNIST &   MLP-1 &       Data Augmentation &          .0065 \\
         F-MNIST &   MLP-1 & Dropout \& Data Augmentation &           .029 \\
         F-MNIST &   MLP-1 &       Batch Normalisation &         .00097 \\
        CIFAR-10 &   CNN-2 &       Dropout &            .38 \\
        CIFAR-10 & CNN-2 &       Data Augmentation &            .13 \\
        CIFAR-10 & CNN-2 & Dropout \& Data Augmentation &            .44 \\
        CIFAR-10 & CNN-2 &        Batch Normalisation &             1.7 \\
    \bottomrule
    \end{tabular}
    \end{small}
    \end{table}

%\begin{figure}[ht]
%\centering
%\includegraphics[width=\linewidth]{images/results_3/d_dist_id=6881.pdf}
%\label{fig:result-4-dist}
%\caption{Histogram of the divergences of $\mathrm{d_u}$ and $\mathrm{d_r}$}
%\end{figure}

%\sat{the results on the scores are really weird and all over the place! someone might read this, as is this a problem of M or the self-influence score? it really hurts your M score i think!}

%\begin{figure}
%  \includegraphics[width=\linewidth]{images/mnist_and_artifact.png}
%  \caption{MNIST image with artifact.}
%  \label{fig:mnist_image_with_artifact}
%\end{figure}

\subsection{Characterisation of unique feature memorisation}\label{sec:discussion}

Our results show that neural networks memorise unique features that occur once in a training dataset, and that regularisation strategies do not help to reduce this behaviour. These results are intriguing and lead us to consider the following questions to characterise unique feature memorisation.

\textit{Is memorisation influenced by sample complexity?} We find a weak correlation between self-influence and $M$ for the top-15, bottom-15 $M$ score canaries in MNIST for MLP-1 (Pearson's correlation coefficient 0.43). And we find no correlation for F-MNIST and CIFAR-10. These results indicate that sample complexity (alone) does not provide an adequate explanation for unique feature memorisation. Next, we explore other possible factors affecting memorisation, such as the stochasticity of the learning algorithm or changes in sample self-influence after a feature has been added.

\begin{figure}[t]
\centering
\subfigure[Variation in the $M$ score for MNIST canaries due to random training seed.]{\label{fig:discussion-a}\includegraphics[width=0.32\linewidth]{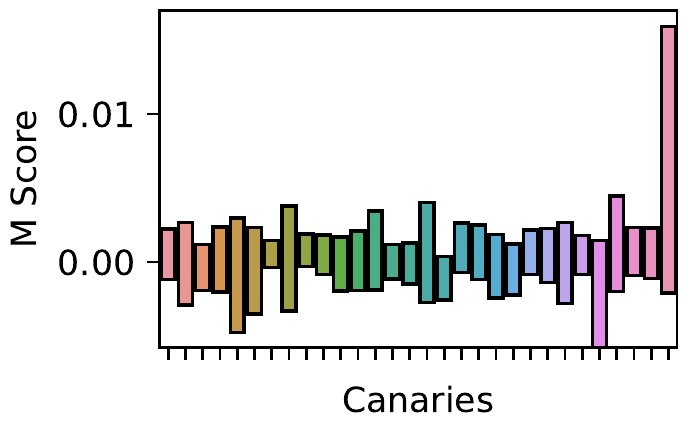}}\hfill
\subfigure[Average activations for the unique/random feature in the second-last layer of MLP-1.]{\label{fig:discussion_activations}\includegraphics[width=0.32\linewidth]{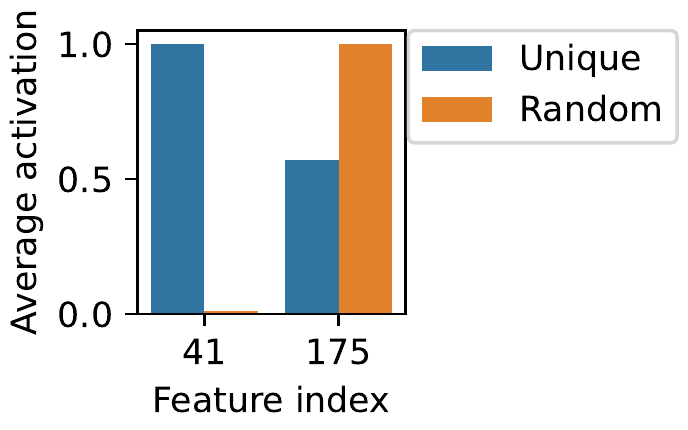}}\hfill
\subfigure[$M$ score profile of an MNIST canary during training.]{\label{fig:hockey_epoch_mnist_mlp-profile}\includegraphics[width=0.32\linewidth]{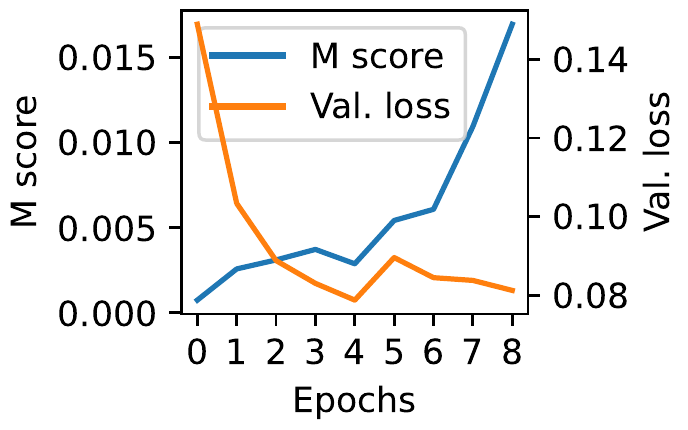}}
\caption{Characterisations of unique feature memorisation in MLP-1 trained on MNIST.}
\label{fig:discussion}
\end{figure}

\textit{Is memorisation affected by training stochasticity?} Yes, we find that unique feature memorisation is affected by the training seed. We use the white box setting to isolate this result from the $M$ score. Figure \ref{fig:discussion-a} shows the range of memorisation scores for 30 randomly selected MNIST canaries over training runs with different random seeds. (Results for 100 canaries in MNIST and CIFAR-10 are given in Section \ref{sec:app-seed}. Positive and negative $M_w$ scores show the randomness in memorisation. 
%This is a worrying sign that model designers cannot control memorisation.

%Crucially $M>0$ and $M\le0$ for the same canary indicates that unique feature memorisation is a random process. Fortunately, the $M$ score allows us to measure the degree of feature memorisation even though its occurrence appears random. 

\textit{Does having a unique feature make for a high self-influence sample?} We  examine the self-influence of samples with and without the unique feature. First, we measure the self-influence of all samples in MNIST using TracIn. Then, we select the highest and lowest influence sample and inject the unique feature. Then, we re-train the model and measure the self-influence of these canaries using the same seed. We find that the influence of the low influence sample increases   hugely from \num{1.0e-13} to \num{7.5e+04}. A high influence sample sees modest increase in influence from \num{6.6e+04} to \num{7.5e+04}. These results show that the addition of the unique feature creates a highly influential sample even when the sample’s other features were easy to learn from other samples. 
%This may have a positive implication in settings where training data are accessible where designers could use self-influence scores to detect leakage.

\textit{Where does memorisation happen in neural networks?} \cite{Baldock2021} showed that sample memorisation occurs deep in neural networks. Here in a white box setting (i.e.\ access to model activations) we show that MLPs extract a latent representation for the unique feature in the layer before the softmax. We describe how we find this latent in Section \ref{sec:add-exp}. Figure \ref{fig:discussion_activations} shows activations for inferences on $\zu$ (unique) and $\zr$ (random). Crucially, only the unique feature activates the latent at index 41. This latent is extracted only for the unique feature and not the random feature $\zr$. Instead, the latent at index 175 is activated by both patches suggesting that this feature represents activations in the location of the patch in the image space.  

\textit{Do unique features get memorised early on?} Figure \ref{fig:hockey_epoch_mnist_mlp-profile} offers a profile of $M$ during training. In agreement with \cite{carlini2019secret, van2021memorization} our results show that unique feature memorisation begins early, and therefore that early stopping does not prevent memorisation.

\textit{Why are unique features memorised?}
% jh
Collectively our results show that unique feature memorisation is not a property of the underlying sample complexity, and that learning on this sample is greatly influenced by the addition of a unique feature. Typically, we would expect the learning algorithm to ignore the unique feature. This is because  under the information bottleneck (IB) principle, information learnt from the other samples in the training dataset is sufficient to reduce the uncertainty on the label distribution \citep{dl-and-information-bottleneck-2015, Achille2018}.

However, in practice this does not seem to be the case. We suggest the following explanation for this behaviour. Let us assume that the classifier is extracting a latent space from the input data.
We can theoretically partition the latents in two parts: those learned from the samples according to the IB principle and those attributed to the unique feature. Indeed our results in Figure \ref{fig:discussion_activations} hint at this. On the latents extracted from the unique feature the classification task is easy: the canary is linearly separable from all samples as these latents are not shared by any other sample. Under the \textit{Principle of Least Effort}  \citep{Geirhos2020-shortcut-learning}, the learning algorithm may shortcut over the unique feature since it is easy to learn, 
%(as the unique feature latents are distinct from the other latents in the sample), 
and as our results suggest it may do so early on in training (Figure \ref{fig:hockey_epoch_mnist_mlp-profile}).
The stochasticity observed in our findings (Figure \ref{fig:discussion-a}), suggests that the decision as to whether the shortcut is learnt is a random process and may relate to the initialisation weights or when the canary is presented to the learning algorithm in stochastic mini-batch gradient descent or other gradient dynamics effects \citep{pezeshki2020gradient}.

%We suggest that some random seeds make it easier for short-cut learning to happen \citep{Geirhos2020-shortcut-learning, }. We suggest that unique feature maybe memorised under the \textit{Principle of Least Effort} when it provides an easy short-cut discriminate between classes. The feature is easy to learn since it has 

%The unique feature can be thought of as a generating factor whose pixel space representation has been removed artificially for all but one image. Once extracted the factor provides an easy point of comparison since it is not present in any other image.

%A unique feature creates a new factor in the dataset which is through the is simply easier to lear

%One possibility is the influence of short-cut learning. For example, under the \textit{Principle of Least Effort}, the network may find it easier to fit the unique feature, than to learn features for training examples from Fashion-MNIST or CIFAR-10 \citep{Geirhos2020-shortcut-learning, pezeshki2020gradient}. For MNIST the task of learning is significantly easier and the unique feature is only memorised on very high influence examples. Another possibility is that TracIn does not trace examples that are necessarily memorable in datasets with more complex features, and hence the unique features are not memorised either.

\section{Related Work}

To the best our of knowledge, no previous study has investigated methods that detect memorisation of single unique features in image classification models in the stringent settings we consider. However, existing research recognises the critical role that memorisation plays in the overfitting and generalisation of neural networks and the security implications of attacks using adversarial training examples. Below we outline work that has provided encouraging findings to help us shape this work.

\subsection{Memorisation}
The main objective of this work is to understand whether memorisation occurs for unique features that occur \textit{once} per dataset and are located on a single training example.

Previous work has established that over-trained, over-parameterised neural networks memorise training labels \citep{Zhang2021-understanding-mem}. Several metrics for assessing the memorisation of training labels have been introduced \citep{feldman-what-neural-networks, Jiang2021}. It was shown that DNNs first learn common patterns in training examples, after which they memorise labels \citep{Arplt2017-a-closer-look-at-memorization, Kim2018a-gan-mem}. More recently, it has been shown that learning and memorisation occur simultaneously \citep{Liu2021}. These works show that regularisation does not eliminate memorisation. We take inspiration from \cite{Zhang2021-understanding-mem} and explore whether the regularisation prevents unique feature memorisation.

% Current privacy attacks
Privacy attacks are methods that exploit data leakage to uncover information about training data. For example, membership inference attacks  deduce whether a sample is in the training set by exploiting a model's overconfidence on examples it has seen \citep{shokri2017membership, Sablayrolles2018, salem2018-ml-leaks, liu2020-have-you-forgotten, choquettechoo2021-label-only-mia}. We propose that models leak information about unique features. And we focus on the memorisation of unique features and not whole data.

%Feature leakage occurs in large language models (LLM) due to unintended memorisation \citep{carlini2019secret, Carlini2020}. Unique sequences, such as credit card numbers or rarely occurring phrases, are random and therefore they are likely to be memorised by a model since they cannot be \textit{learnt} from similar patterns in the remainder of the training data.

Recent work by \citet{carlini2019secret, Carlini2020} established that LLMs memorise features.  Unique sequences, such as credit card numbers or rarely occurring phrases are random and therefore cannot be \textit{learnt} from similar patterns in the remainder of the training data. The authors developed \textit{exposure} as a black box inference method to measure feature memorisation. This work has inspired us to draw upon the use of canaries to measure the memorisation of unique features in images. We follow their idea to also explore memorisation before \textit{over-learning} occurs. However, our work is different to measuring feature memorisation in LLMs. In discriminative models we cannot measure the propensity of a unique feature, and other theoretically equally likely features from the softmax outputs, and thus we cannot calculate the exposure of the canary. Also in LLMs the feature's label is self-evident, whereas the under the same assumption we do not know the unique feature's label in the discriminate setting. This makes memorisation harder to measure.

Property inference attacks attempt to learn a group property/feature of the dataset. For example, what proportion of people in the training set wear glasses? \citep{ateniese2013-property-inference, ganju2018-property-inference}. These attacks are typically white box and proceed by using a shadow model to make inferences on the target model weights. Feature memorisation, as we investigate here, can be viewed as an extreme property inference attack where a unique feature, a person who wears glasses, occurs only once in the dataset. Existing approaches, however, cannot address unique feature memorisation since labelling the training weights to train the shadow model requires ground-truth knowledge of whether the feature was memorised or not.

We also draw upon the idea of concept activation vectors (CAV), as introduced by \citet{kim2018interpretability-cav}, to infer the sensitivity of the network to the feature we are interested in vs.\ random features that we are not. However, instead of image concepts such as stripes, we focus on unique features and use only the outputs of the trained model as opposed to the internal activations of the network. Our memorisation score also does not require the label of the unique feature.

A concurrent work by \cite{Understanding_Rare_Spurious_Correlations_in_Neural_Networks_2022} show in a white box setting that neural network image classifiers learn rare features when they occur \textit{three or more} times in the training data. We instead develop a method which does not require access to training data and show that  memorisation occurs even for a \textit{single} unique feature in the training data.

\subsection{Backdoor attacks}

%In a backdoor attack an adversary alters a model's test-time behaviour on images which contain a key or feature generated with an optimisation algorithm by poisoning several mislabelled training images with the key \cite{chenLiu2017, guBadNets2017, Liu2018TrojaningAO, munoz2017towards}. Similarly to our work, \citet{cleanlabelbackdoor2018, hiddenbackdoorattacks2019}, perform attacks using clean labelled data. However, the key or adversarial image is generated with an optimisation algorithm. Our work differs from these works since we do not use optimised unique features, we introduce only a single unique feature to the training data, and we train models end-to-end.

Backdoor attacks are fundamentally different to our work. These attacks attempt to adversarially change a model's predictions by injecting an optimised image patch onto training examples such that when this patch occurs on an attack example at test time, the predictions of the model can be controlled \citep{chenLiu2017, guBadNets2017, Liu2018TrojaningAO, munoz2017towards, cleanlabelbackdoor2018, hiddenbackdoorattacks2019}. In contrast, we show that a unique feature which occurs in the training data is memorised. This feature is not optimised to modify the outputs of the model at test time.

\section{Conclusion}

We present a score to measure the memorisation of unique features in imaging datasets by neural network image classification models. We focus on the case where the unique feature appears on a single image in the training data, and where we have access to the the unique feature, but not to the training data or the model weights.

We show that unique feature memorisation occurs in this setting, and is not eliminated by typical explicit and implicit regularisation strategies, dropout, data augmentation and batch normalisation. We derive these results for benchmark datasets and a range of neural network architectures. We also show that unique feature memorisation occurs early in training, and that MLPs extract representations for theses features.

\textbf{Social impact} These results, even in standard benchmark datasets, suggest that neural networks pose a privacy risk to unique sensitive information in imaging datasets even if the information occurs once. The information does not have to be rare in the wild. In the context of a healthcare application, the information could be a patient name that was not removed by an image pre-processor.

\section{Acknowledgements}
%\sat{dont forget to hide ack and names!}
This work is supported by iCAIRD, which is funded by Innovate UK on behalf of UK Research and Innovation (UKRI) [project number 104690]. S.A.\ Tsaftaris acknowledges also support by a Canon Medical / Royal Academy of Engineering Research Chair under Grant RCSRF1819\textbackslash8\textbackslash25. This work
was partially supported by the Alan Turing Institute under EPSRC grant EP/N510129/1.

\bibliography{main.bib}
\bibliographystyle{apalike}

\appendix

\section{Models and training}\label{sec:models-and-training}

\subsection{Datasets}

We choose three benchmarked datasets to train the image classification models. MNIST is a 10 class set of handwritten digits with image size $28\times28\times1$ \citep{lecun2012mnist}, F-MNIST is a 10 class set clothing thumbnails with image size $28\times28\times1$ \citep{xiao2017fashion} and CIFAR-10 is a 10 class set of common objects with image size $32\times32\times3$ \citep{krizhevsky2009-cifar10}. We use the original train/test splits given by the dataset authors.

We use CIFAR-10 as an OOD dataset to evaluate the M-score for models trained on MNIST and F-MNIST, and to evaluate models trained on CIFAR-10, we use a resized, 3-channel version of MNIST. To perform image transformations we use the \texttt{tf.image} API in TensorFlow v2.7.

\subsection{Training strategy}

All models are trained to reduce overlearning. To do this we make use of using early stopping. We train up to 500 epochs with a patience of 10 epochs. After training, we select the final model weights from the epoch with the lowest validation loss. We use the Adam optimiser and a cross-entropy loss function \citep{kingma2017adam}.

To train and evaluate models we use TensorFlow 2.7. We train models using an Nvidia\textsuperscript{\textregistered} Titan RTX\texttrademark. A low-carbon and renewable energy source was provided by The University of Edinburgh. We estimate the computation time for the experiments to be around 200 GPU hours.

\subsection{Network architectures}

We evaluate our memorisation score using several common architectural styles of neural networks. The first, \textit{MLP-1} is trained on MNIST and Fashion-MNIST datasets. It is comprised of: \textrm{Dense(512)} $\rightarrow$ \textrm{ReLU} $\rightarrow$ \textrm{Dense(256)} $\rightarrow$ \textrm{ReLU} $\rightarrow$ \textrm{Dense(128)} $\rightarrow$ \textrm{ReLU} $\rightarrow$ \textrm{Dense(\#classes)} $\rightarrow$ \textrm{Softmax}.

We train MLP-1 with a learning rate of \num{3e-4} and a batch size of 128.

CNN-1 a simple 2-layer convolutional neural network. It is comprised of: \textrm{Conv2D(32,3,3)} $\rightarrow$ \textrm{ReLU} $\rightarrow$ \textrm{Conv2D(64,3,3)} $\rightarrow$ \textrm{MaxPool2d(2,2)} $\rightarrow$ \textrm{ReLU} $\rightarrow$ \textrm{Dense(128)} $\rightarrow$ \textrm{ReLU} $\rightarrow$ \textrm{Dense(128)} $\rightarrow$ \textrm{ReLU} $\rightarrow$ \textrm{Dense(\#classes)} $\rightarrow$ \textrm{Softmax}.

We train CNN-1 with a learning rate of $\num{3e-4}$ and a batch size of 128.

\textit{CNN-2} is small VGG-style network trained on CIFAR-10. It is comprised of: \textrm{Conv2D(32,3,3)} $\rightarrow$ \textrm{ReLU} $\rightarrow$ \textrm{Conv2D(32,3,3)} $\rightarrow$ \textrm{ReLU} $\rightarrow$ \textrm{MaxPool2d(2,2)} $\rightarrow$ \textrm{Conv2D(64,3,3)} $\rightarrow$ \textrm{ReLU} $\rightarrow$ \textrm{Conv2D(64,3,3)} $\rightarrow$ \textrm{ReLU} $\rightarrow$ \textrm{MaxPool2d(2,2)} $\rightarrow$ \textrm{Dense(1024)} $\rightarrow$ \textrm{ReLU} $\rightarrow$ \textrm{Dense(\#classes)} $\rightarrow$ \textrm{Softmax}.

We train CNN-2 with a learning rate of \num{3e-4} and a batch size of 512.

\textit{DenseNet} is DenseNet trained on CIFAR-10 \cite{huang2018densely}. The network has 100 layers, a growth factor of 12, and three dense blocks. We use an existing implementation and train using the same parameters given in \citet{Atienza}.

\subsection{Regularisation strategies}

We sample new data augmentations for every mini-batch that we train. For MNIST and F-MNIST three augmentations. We randomly adjust the contrast by a factor of 0.2 and we apply random cropping (27, 27) so that we ensure we do not remove the unique features from the image. For CIFAR-10 we also include random horizontal flipping.

We use dropout in MLPs and CNNs. For each mini-batch we reduce the connections in fully connected layers by a factor of 0.2, and in CNN layers we reduce by a factor of 0.5. No dropout is applied to the softmax layer. Batch normalisation is applied directly before the activation layer in all layers in MLP-1 and the CNNs except for the softmax layer. We use $\textrm{momentum}=0.99$ and $\epsilon=0.001$.

\subsection{Computing self-influence using TracIn}

We determine the self-influence of samples using TracIn. TracIn measures the self-influence of an example by computing the difference in training loss with respect to itself between successive iterations of stochastic gradient descent (SGD) where the loss is computed on that example. We use an approximation, TracInCP, which approximates TracIn for SGD with mini-batches. It is computed at user specified checkpoints given by

\begin{equation}
    \textrm{TracInCP} = \sum_i^n {\eta_i} ||\nabla l (\vec{w}_i, \vec{x})||^2,
    \label{tracincp}
\end{equation}

where $\vec{w}_i$ and $\eta_i$ are the weights and learning rate of $f$ and the optimisation algorithm at checkpoint $i$ to $n$, $\vec{x}$ image whose self influence we want to measure.

% using tracin
For each black box experiment in this work, we measure the self-influence of all training examples by selecting 10 evenly spaced checkpoints which account for a 95\% reduction in the training loss \citep{Garima-2022-faq}. We select the top-15 and bottom-15 examples as canaries. We create 30 models, each of which is trained on the canary that has been injected with a unique feature.

\subsection{Finding a latent represent of the unique feature in an MLP}

We train 100 MLP models with a fixed seed using a randomly selected canary per model. We perform inferences using the training data containing the unique feature and the clean OOD image pairs, and record the activations in the second-last layer. Next, we threshold the activations such any non-zero activation is equal to one. Since ReLU activations are used, the remaining activations are zero. We subtract the clean activations from the unique feature activations and repeat the thresholding operation. In this way we measure activations only excited by the unique feature patch. Next, we average over the activations map for each model and normalise. We obtain the most commonly excited activations marginalised over the choice of canary.

\section{Additional experiments}\label{sec:add-exp}

\subsection{Memorisation in a white box setting}\label{app:greybox}

\begin{figure}[!htb]
\centering
\subfigure[M Score for 100 randomly chosen MNIST samples used to train MLP-1.]{\label{fig:milkweed-m1-sub1}\includegraphics[width=0.45\linewidth]{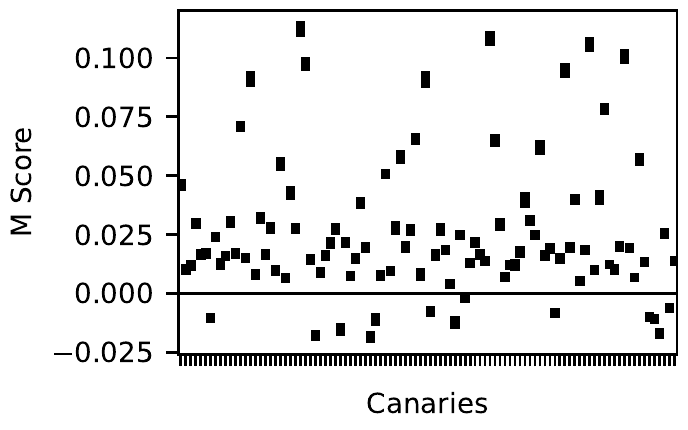}}
\hfill
\subfigure[M Score for 100 randomly chosen CIFAR-10 samples used to train CNN-2.]{\label{fig:milkweed-m1-sub2}\includegraphics[width=0.45\linewidth]{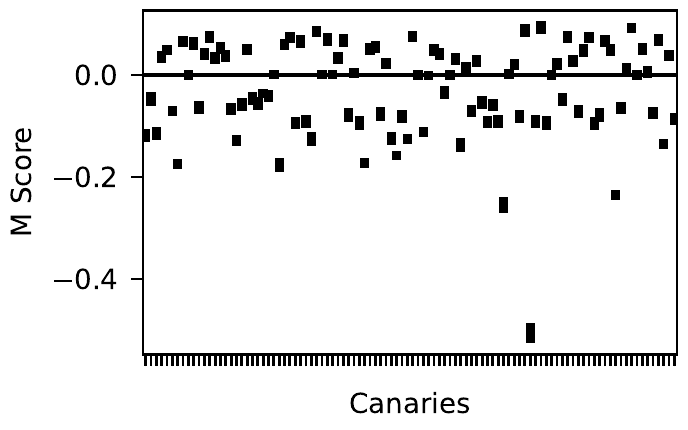}}
\caption{M scores for the unique feature patch on 100 randomly chosen canarys. Each datapoint represents the M score for a single canary in a training dataset used to train a single model. We show the standard deviation of the M score of three evaluations.}
\label{fig:greybox-ablation-m}
\end{figure}

In this experiment we show the memorisation scores for 100 randomly selected MNIST and CIFAR-10 canaries for MLP-1 and CNN-2 respectively. We measure the memorisation scores for three evaluations of $M_w$. In each evaluation we sample a different set of random features to inject into the OOD examples, and we use all the samples in the OOD dataset. Each evaluation acts on the same trained model. In Figure \ref{fig:greybox-ablation-m} for we show the results for the mean $M_w$ score and one standard deviation.

\subsection{Effect of learning algorithm on memorisation}\label{sec:app-seed}

\begin{figure}
    \centering
    \includegraphics{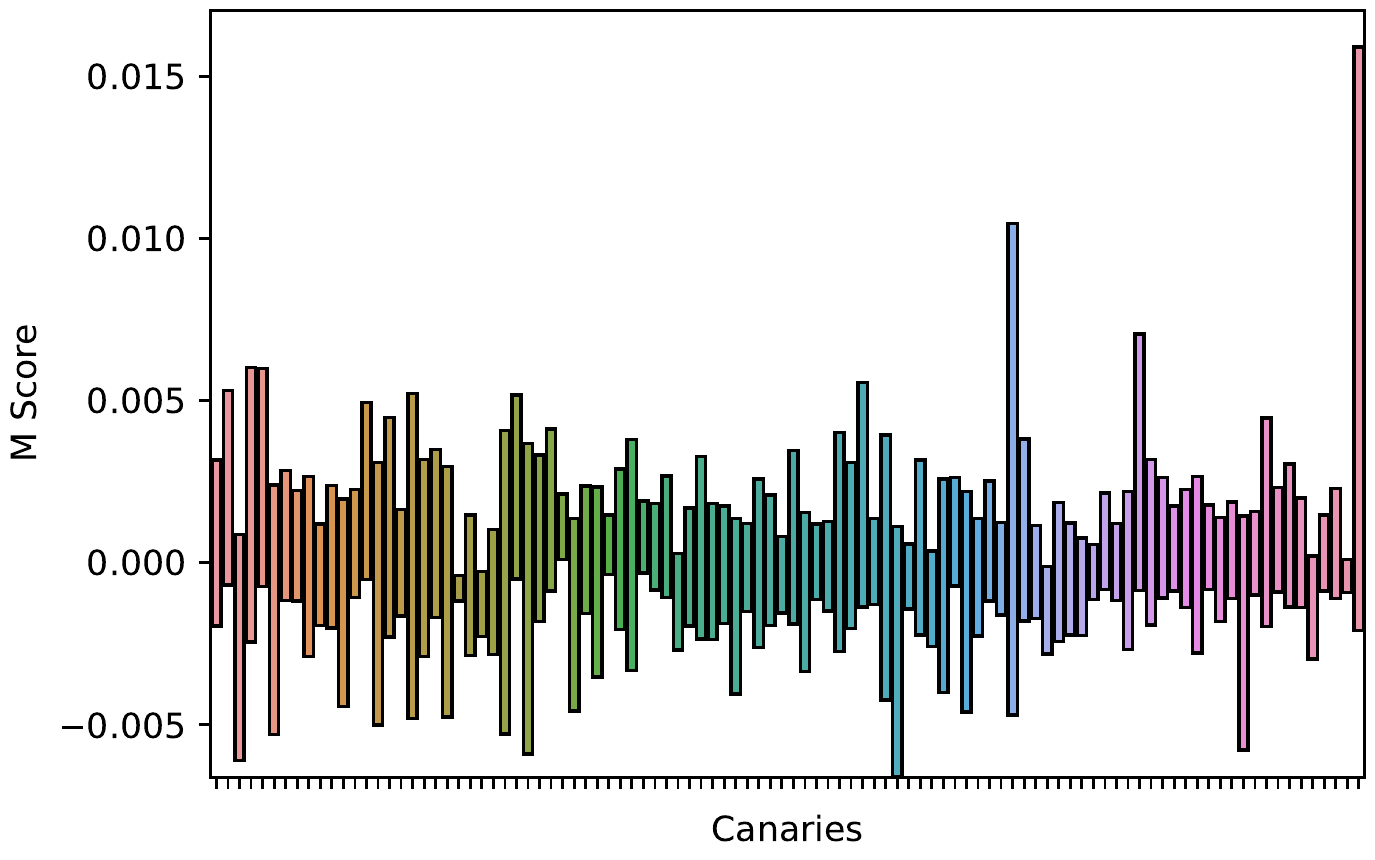}
    \caption{Range in the $M$ score for MNIST canaries across five training runs.}
    \label{fig:m-score-var-full}
\end{figure}

\begin{figure}
    \centering
    \includegraphics{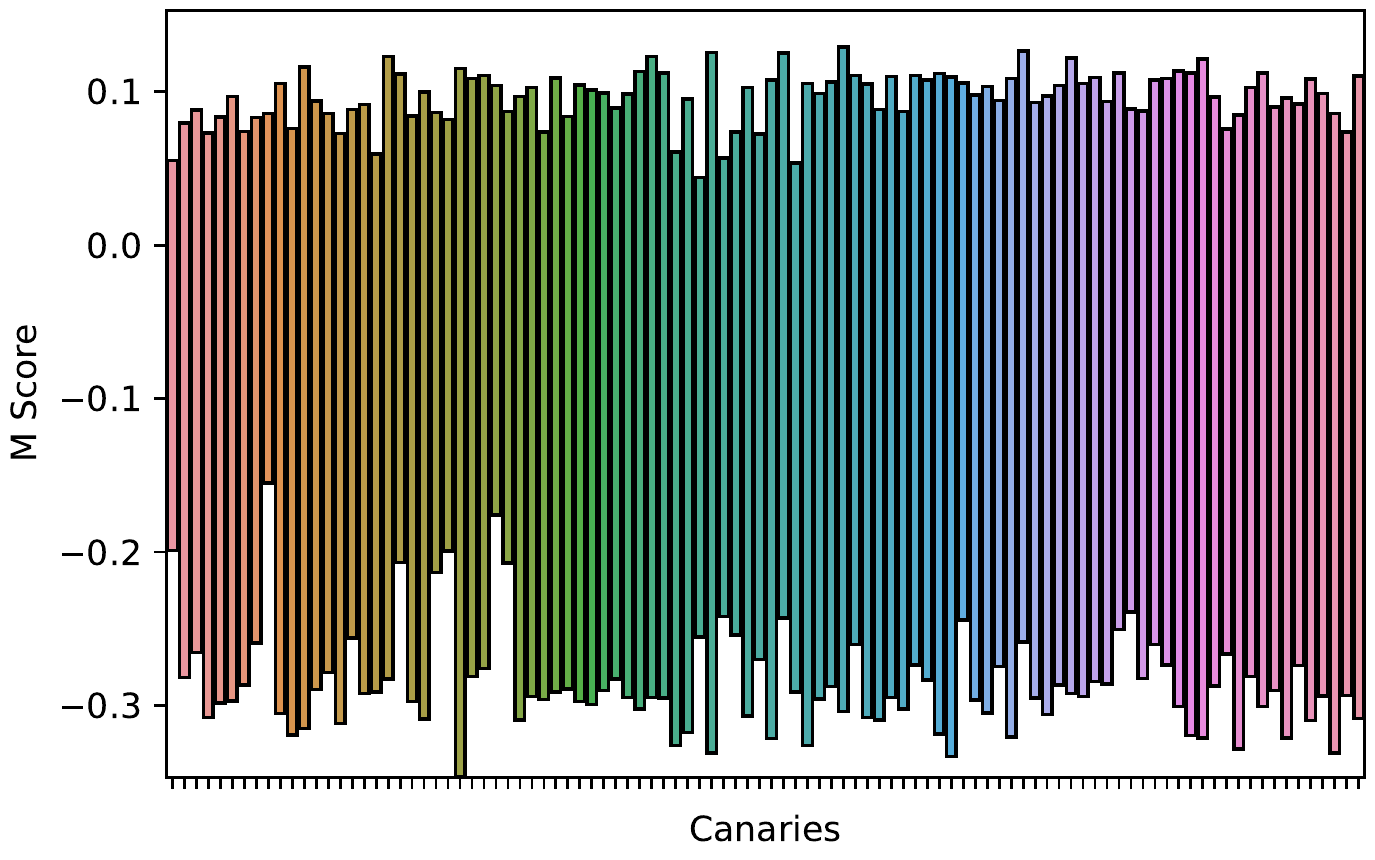}
    \caption{Range in the $M$ score for CIFAR-10 canaries across five training runs.}
    \label{fig:m-score-var-full-cifar-10}
\end{figure}

In this experiment we measure $M$ for 100 randomly selected canaries over five training runs. We train 100 models, one for each canary, and we repeat the training five times using difference random seeds. The $M$ score uses the same random seed for each evaluation. We train MLP-1 on MNIST and CNN-2 on CIFAR-10. We calculate the range of $M$ over the training runs. The results are shown in Figure \ref{fig:m-score-var-full} and \ref{fig:m-score-var-full-cifar-10} respectively. Both figures show that $M$ has positive and negative results in different training runs for the same canaries. This shows that memorisation varies depending on the randomness of the learning algorithm. Also the range of the $M$ score over training runs is far greater than the range due to the stochasticity of the memorisation score in Section \ref{app:greybox}. This shows that the variation due to the learning algorithm is greater than the precision of the memorisation score.

%{\section{Variation in M score for CIFAR-10 canaries due to training seed}

%\begin{figure}[htp]
%    \centering
%    \includegraphics{images/milkweed_c1/milkweed_c1_cifar_cnn.pdf}
%    \caption{Variation in the M score for CIFAR-10 canaries due to random training seed. \textcolor{purple}{Full experiment is running}}
%    \label{fig:milkweed_c1_cifar_cnn}
%\end{figure}}

\end{document}